\documentclass[]{config/antgroup}
\usepackage{config/antgroup}
\newcommand*\justify{%
  \fontdimen2\font=0.4em% interword space
  \fontdimen3\font=0.2em% interword stretch
  \fontdimen4\font=0.1em% interword shrink
  \fontdimen7\font=0.1em% extra space
  \hyphenchar\font=`\-% allowing hyphenation
}

\renewcommand{\texttt}[1]{%
  \begingroup
  \ttfamily
  \begingroup\lccode`~=`/\lowercase{\endgroup\def~}{/\discretionary{}{}{}}%
  \begingroup\lccode`~=`[\lowercase{\endgroup\def~}{[\discretionary{}{}{}}%
  \begingroup\lccode`~=`.\lowercase{\endgroup\def~}{.\discretionary{}{}{}}%
  \catcode`/=\active\catcode`[=\active\catcode`.=\active
  \justify\scantokens{#1\noexpand}%
  \endgroup
}

\PassOptionsToPackage{numbers, compress}{natbib}

\usepackage{graphicx}
\usepackage{todonotes}
\usepackage{multirow}
\usepackage{hyperref}
\usepackage[capitalize,noabbrev,nameinlink,sort]{cleveref}
\usepackage{subcaption}
\usepackage{multicol}
\usepackage{array}
\usepackage{enumitem}
\usepackage{algorithm}
\usepackage{algpseudocode}
\usepackage{tabularx}
\usepackage{makecell}
\usepackage{xspace}
\usepackage{colortbl}
\usepackage{fontawesome5}
\usepackage{simpleicons}
\usepackage{pifont}  
\usepackage[normalem]{ulem}
\useunder{\uline}{\ul}{}
\usepackage{tcolorbox}
\usepackage{tablefootnote}

\usepackage{amsmath}
\usepackage{amssymb}
\usepackage{amsfonts}                               % blackboard math symbols
\usepackage{amsthm}
\usepackage[mathcal]{eucal}
\usepackage{mathrsfs}
\usepackage{bm}                                     % bm command
\usepackage{blkarray}                               % to support matrix
\usepackage{nicefrac}                               % compact symbols for 1/2, etc.
\usepackage{bbm}

\usepackage{wrapfig}
\usepackage{graphicx}                               % include pdf figures
\usepackage{caption}
\usepackage{cleveref}
\usepackage{tikz}                                          
\usepackage{circuitikz}
\usetikzlibrary{patterns,snakes}
\usetikzlibrary{positioning,calc,fit,decorations.pathmorphing,shapes.geometric, shapes.gates.logic.US, calc}
\usetikzlibrary{arrows,arrows.meta,decorations.markings,shapes,shapes.arrows}
\usetikzlibrary{decorations,decorations.pathreplacing}
\usetikzlibrary{backgrounds}
\usepackage{filecontents}                           % support to pgfplots
\usepackage{pgfplots}
\usepackage{pgfplotstable}
\usepgfplotslibrary{groupplots}
\usepackage{scalefnt}
\pgfplotsset{compat=newest}
\usepackage{xcolor}

\definecolor{firstcolor}{HTML}{C3423F}
\definecolor{secondcolor}{HTML}{2A4B8C}
\definecolor{aworld_blue}{HTML}{4e81ff}
\definecolor{aworld_cyan}{HTML}{41d7fa}
\definecolor{aworld_teal}{HTML}{5fede4}

\hypersetup{
    colorlinks=true,
    linkbordercolor=aworld_blue,
    citebordercolor=aworld_cyan,
    urlbordercolor=aworld_teal
}

\newcommand{\aworld}{\textsc{AWorld}\xspace}
\def\eqref#1{equation~\ref{#1}}
\def\1{\bm{1}}

\DeclareMathAlphabet{\mathsfit}{\encodingdefault}{\sfdefault}{m}{sl}
\SetMathAlphabet{\mathsfit}{bold}{\encodingdefault}{\sfdefault}{bx}{n}

\title{\raisebox{-0.2em}{\includegraphics[height=1.1em]{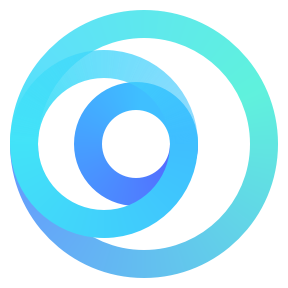}}AWorld: Orchestrating the Training Recipe for Agentic AI}

\author{
  Chengyue Yu$^{1,\ast}$, Siyuan Lu$^{1,2,3,\ast}$, Chenyi Zhuang$^{1,\dagger}$, Dong Wang$^{1}$, Qintong Wu$^{1}$, Zongyue Li$^{1}$,\\
  Runsheng Gan$^{1}$, Chunfeng Wang$^{1}$, Siqi Hou$^{1}$, Gaochi Huang$^{1}$, Wenlong Yan$^{1}$, Lifeng Hong$^{1}$, \\
  Aohui Xue$^{1}$, Yanfeng Wang$^{1}$, Jinjie Gu$^{1}$, David Tsai$^{1}$,  Tao Lin$^{3,1}$
}

{

\footnotetext{$^\ast$Equal contributions. Work was done during Siyuan's internship in Ant Group. }
\footnotetext{$^\dagger$Corresponding Author.}}

\affiliation{$^1$AWorld Team, Inclusion AI\quad}
\affiliation{$^2$Shanghai Innovation Institution\quad}
\affiliation{$^3$Westlake University\\}

\begin{document}

\maketitle

\begin{center}
  \vspace{-1.5em}
  \href{https://github.com/inclusionAI/AWorld/tree/main/train}{\faGithub\ \texttt{https://github.com/inclusionAI/AWorld/tree/main/train}}
  \\
  \vspace{0.25em}
  \href{https://huggingface.co/inclusionAI/Qwen3-32B-AWorld}{%
    \raisebox{-0.1em}{\includegraphics[height=1.0em]{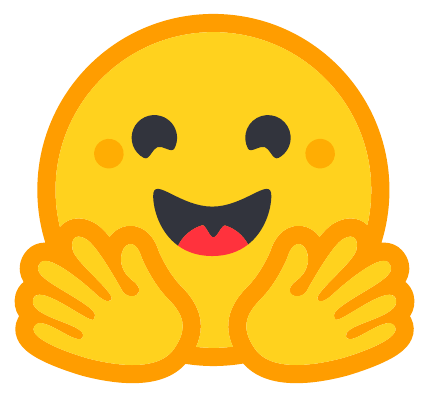}}\
    \texttt{https://huggingface.co/inclusionAI/Qwen3-32B-AWorld}%
  }
  \vspace{0.5em}
\end{center}

%\clearpage
%\tableofcontents
%\newpage
%%%%%%%%%%%%%%%%%%%%%%%%%%%%%%%%%%%%%%%%%%%%%%%%%%%%%%%%%%%%%%%%%%
%%%%%%%%%%%%%%%%%%%%%%%%%%%%%%%%%%%%%%%%%%%%%%%%%%%%%%%%%%%%%%%%%%
%%%%%%%%%%%%%%%%%%%%%%%%%%%%%%%%%%%%%%%%%%%%%%%%%%%%%%%%%%%%%%%%%%
% main document

% 从第二页开始使用带页眉的样式
\begin{abstract}
  The ``learning from practice'' paradigm is crucial for developing capable Agentic AI systems, yet it is severely hampered by inefficient experience generation, a bottleneck especially pronounced in complex benchmarks like GAIA. To address this, we introduce \aworld, an open-source system engineered for large-scale agent-environment interaction. By distributing tasks across a cluster, \aworld accelerates experience collection by $14.6$x compared to standard single-node, sequential execution. This critical speedup makes extensive reinforcement learning practical and scalable. Leveraging this capability, we trained a Qwen3-32B-based agent that achieves pass@1 accuracy of 32.23\% on the GAIA test set, which surpasses GPT-4o (27.91\%) and rivals DeepSeek-V3 (31.89\%). Our open-source system and the resulting agent provide a practical blueprint for a complete agentic AI training pipeline, from efficient interaction to demonstrable model improvement.
\end{abstract}

\begin{figure}[H]
  \centering
  \begin{subfigure}[]{0.4\textwidth}
    \centering
    \includegraphics[width=\linewidth]{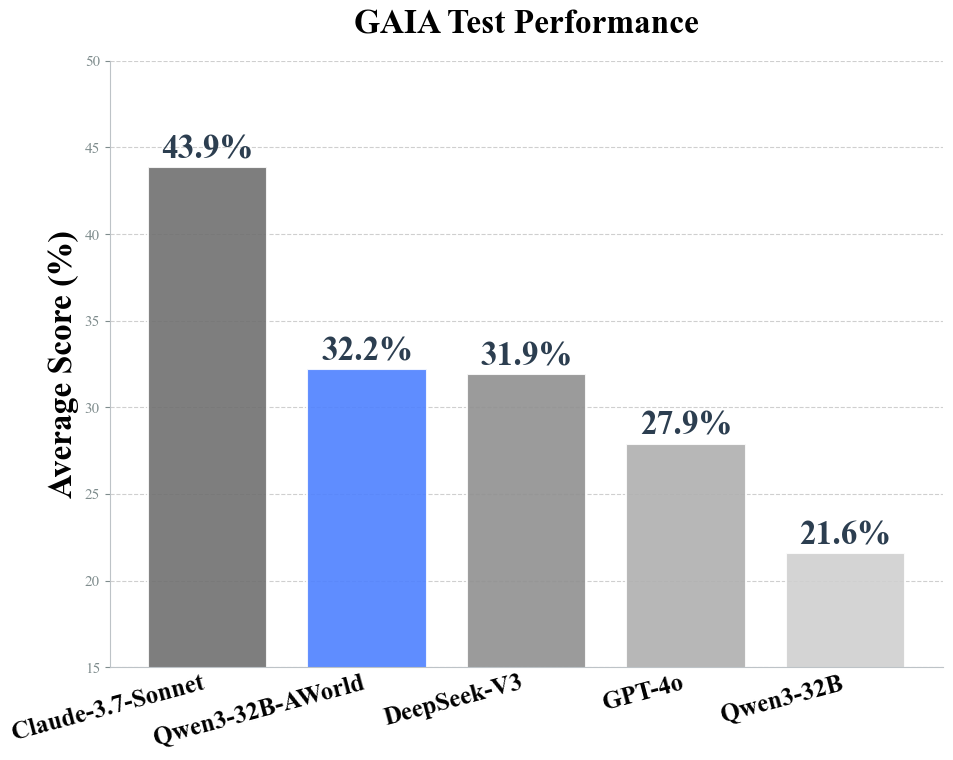}
  \end{subfigure}%
  \begin{subfigure}[]{0.6\textwidth}
    \centering
    \includegraphics[width=\linewidth]{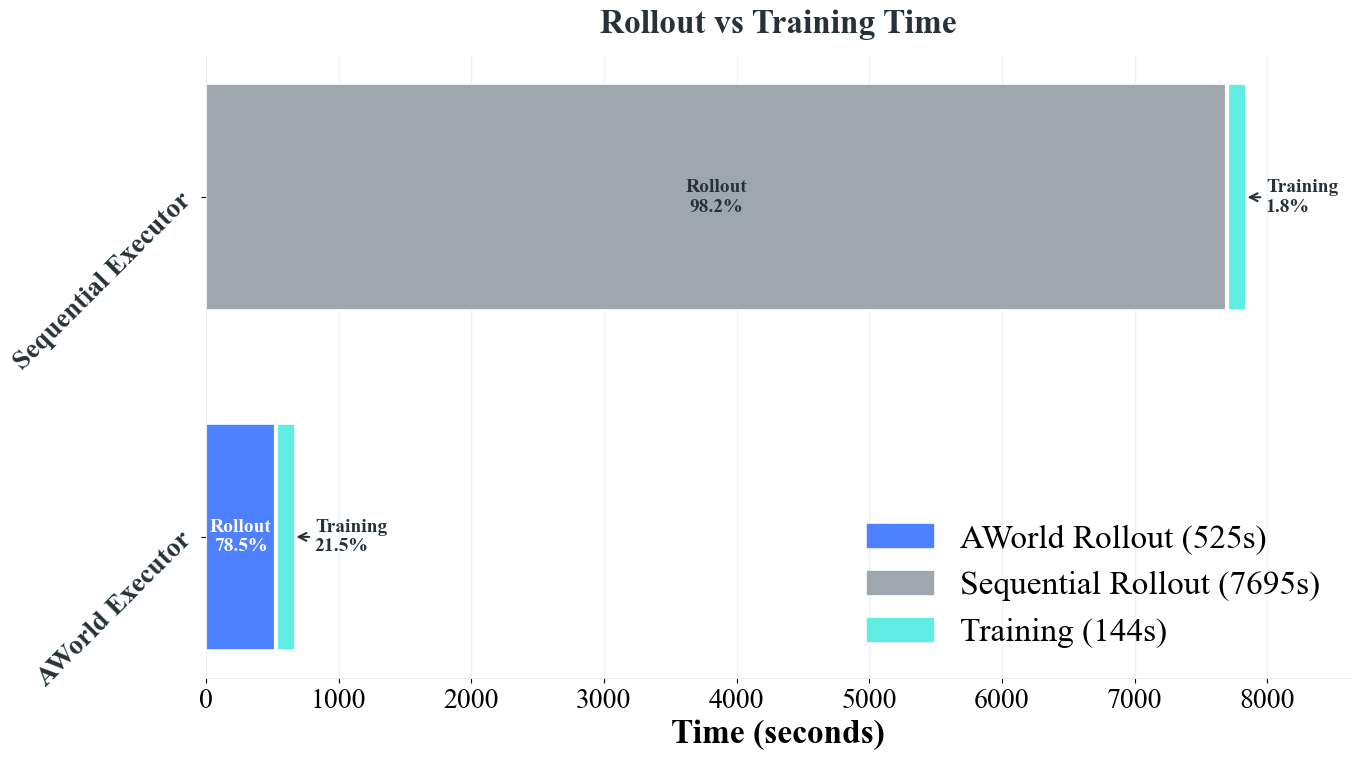}
  \end{subfigure}
  \caption{\textbf{\aworld: High Efficiency Enables High Performance on GAIA.}
  \textbf{(Left)} By using the \aworld framework to conduct fine-tuning and reinforcement learning on the Qwen3-32B base model, our resulting agent (Qwen3-32B-\aworld) demonstrates a substantial performance gain. It achieves a pass@1 score that is highly competitive with frontier proprietary models like GPT-4o.
  \textbf{(Right)} This effective training is made practical by \aworld's core design. Its distributed architecture accelerates the critical experience generation (rollout) phase by a factor of 14.6x, overcoming the primary bottleneck faced by standard single-node, sequential processes.}
  \label{fig:main}
\end{figure}
\newpage
\section{Introduction}
\label{sec:intro}

\begin{figure}[!t]
  \centering
  \includegraphics[width=\textwidth]{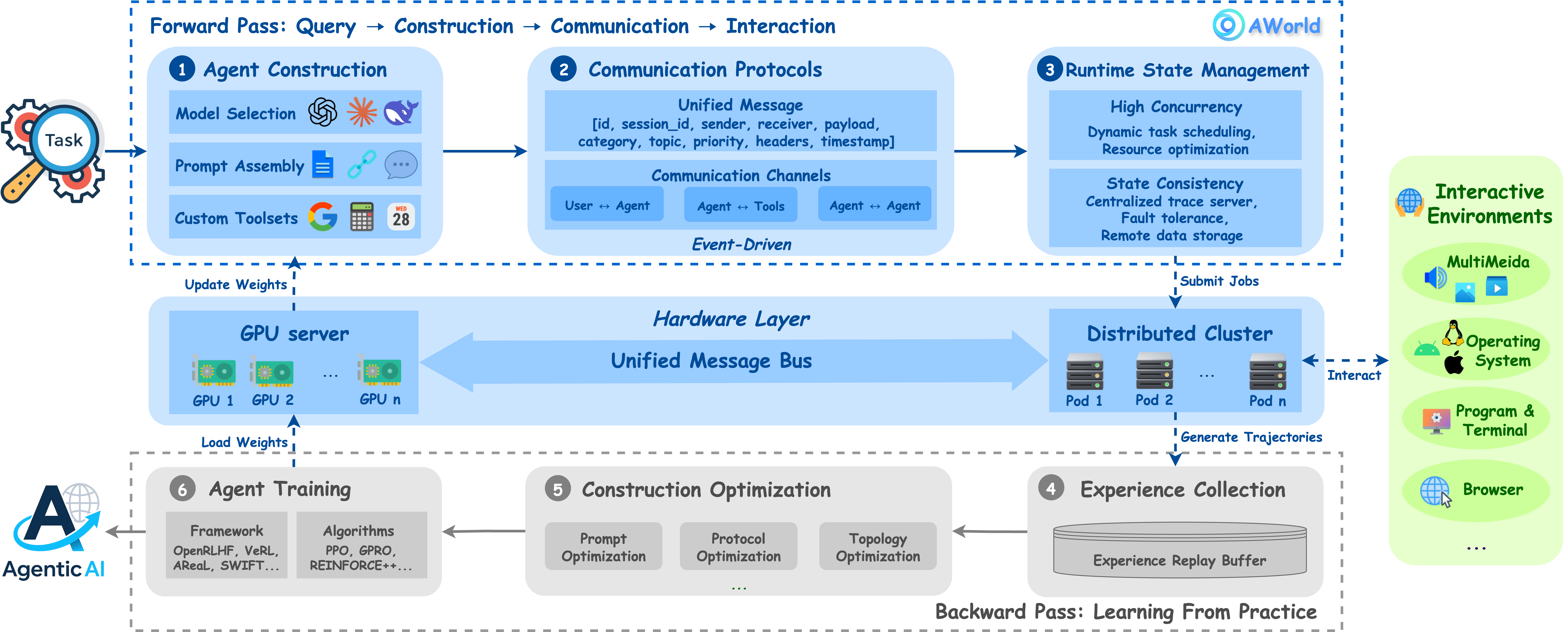}
  \caption{\textbf{The architecture of \aworld, a framework for Agentic AI designed around the ``learning from practice'' paradigm}. The framework operates in two main flows: a \emph{Forward Pass} (top; \ding{172}-\ding{174}), where agents are constructed and interact with complex environments to generate task-solving trajectories; and a \emph{Backward Pass} (bottom; \ding{175}-\ding{177}), where these trajectories are used as experience to train agents and optimize the entire system via reinforcement learning. This closed-loop design enables agents to continuously improve their performance on real-world problems.}
  \label{fig:intro}
\end{figure}

Since the release of ChatGPT~\citep{achiam2023gpt} by OpenAI in 2022, Large Language Models (LLMs)~\citep{anthropic2024claude37sonnet,touvron2023llama,team2023gemini} have demonstrated remarkable capabilities across diverse domains, achieving expert human-level performance in many domains like mathematics~\citep{deepmind2024imo}, demonstrating that AI has reached intellectual capabilities that rival human expertise in numerous domains.
However, despite the impressive capabilities of individual LLMs, current AI systems still struggle to solve real-world, complex tasks effectively. For instance, on the challenging GAIA benchmark~\citep{mialon2023gaia}, even closed-source models like GPT-4~\citep{achiam2023gpt} achieve only 3.99\% accuracy, highlighting the significant gap between LLM capabilities and agent performance in complex, multi-step reasoning scenarios.

The future direction of AI development lies in addressing real-world problems that require complex multi-turn interactions with environments through long trajectories called \emph{Agentic AI}~\citep{wooldridge1995intelligent,sapkota2025ai}. This paradigm emphasizes ``\textbf{learning from practice}'', where agents must continuously engage with external environments to solve complex, multi-step reasoning tasks. As highlighted by recent research~\citep{yao2025secondhalf, silver2025welcome}, the training recipe for agentic AI consists of three fundamental components: (1) \textbf{Algorithm} - the learning mechanisms that enable agents to adapt and improve from environmental interactions; (2) \textbf{Environment} - the complex, interactive settings that provide rich feedback and diverse challenges for agent learning; and (3) \textbf{Priors} - the foundational capabilities of current large models in reasoning, mathematics, vision, and other domains that serve as the starting point for agent specialization.

However, realizing the ``learning from practice'' paradigm faces critical challenges tied directly to these core components. For the Algorithm component, the high cost of curating realistic and complex tasks often leads to data scarcity—for instance, the entire GAIA~\citep{mialon2023gaia} validation set contains only 165 questions. This limited availability of high-quality data underscores the continuous need for more sample-efficient learning methods, as current techniques often require vast amounts of experience to achieve proficiency.
Simultaneously, for the Environment component, while valuable interactive environments for tasks like browser navigation~\citep{zhou2024webarena}, computer control~\citep{xie2024osworld}, and web shopping~\citep{yao2022webshop} have emerged, they remain scarce and often present significant challenges in deployment and scalability. Even with advanced algorithms and rich environments, the sheer volume of interaction required for an agent to gather meaningful experience becomes a daunting logistical hurdle. These challenges converge on a central bottleneck: \emph{the inefficiency of the agent-environment interaction loop}, which is the primary obstacle to building more capable agents.

Facing these multifaceted challenges, the community urgently needs a comprehensive framework that integrates all components required for ``learning from practice'', focusing on end-to-end optimization of the training recipe for agentic AI. To address this need, we propose \aworld, a framework that provides comprehensive support across three critical dimensions:
\begin{enumerate}[leftmargin=12pt, nosep]
  \item \textbf{Prior Model Selection}: \aworld provides a unified interface for model integration, supporting flexible configuration of target models across different training frameworks (detailed in \cref{agent_cons}).
  \item \textbf{Runtime Construction}: Beyond high-concurrency support, \aworld encapsulates communication protocols between models and tools, as well as inter-agent communication protocols (\cref{com_pro}), while implementing robust state management capabilities to handle complex tasks with extended contexts (\cref{dist_env}).
  \item \textbf{RL Algorithm Design}: Although \aworld is not itself a training framework, it seamlessly integrates with reinforcement learning frameworks including OpenRLHF~\citep{hu2024openrlhf}, VeRL~\citep{sheng2025hybridflow}, AReaL~\citep{fu2025areal}, and SWIFT~\citep{zhao2025swift}, thereby unifying model training clusters with environment inference systems. Details of this integration are discussed in~\cref{subsec:training_orchestration}.
    Collectively, these foundational features position \aworld as an ideal platform for implementing \textbf{End-to-End learning-from-practice} pipelines, as demonstrated in~\cref{fig:intro}.
\end{enumerate}

Overall, in this paper, we have made the following contributions:
\begin{itemize}[leftmargin=12pt, nosep]
  \item \textbf{We design and implement \aworld, a modular and scalable open-source framework} that serves as the core infrastructure for the entire ``learning from practice'' lifecycle of AI agents. \aworld provides unified solutions for agent construction, communication, distributed execution, and training orchestration to tackle complex, long-horizon tasks.
    \looseness=-1

  \item \textbf{We conduct a systematic analysis on the GAIA benchmark to empirically demonstrate that agent performance is critically bottlenecked by the efficiency of experience generation (rollouts).} We then show that \aworld's distributed architecture resolves this bottleneck, achieving a significant speedup in data collection compared to a standard sequential approach, making large-scale agent training feasible.

  \item \textbf{By leveraging the \aworld framework, we successfully train an open-source agent based on Qwen3-32B.} This agent not only significantly surpasses its base model but also achieves highly competitive performance on the challenging GAIA benchmark, even outperforming strong proprietary models like Claude-3.7-Sonnet on the highest difficulty levels.
\end{itemize}
\section{\aworld Framework}
\label{intro_framework}

In this section, we present the \aworld framework, which serves as a general-purpose infrastructure for building, deploying, and training intelligent agents in complex environments.
We introduce the four key components of \aworld:
\begin{itemize}[leftmargin=12pt, nosep]
  \item \textbf{Agent Construction:} We begin at the foundational level by simplifying the instantiation of an individual agent, defining its core logic, toolset, and planning capabilities.
  \item \textbf{Communication Protocols:} With a single agent defined, we then establish a unified message-passing architecture, enabling it to interact reliably with itself, other agents, various tools and environments.
  \item \textbf{Runtime State Management:} To scale these interactions for complex tasks, the next logical layer provides robust distributed execution, managing the state of numerous concurrent agents across a cluster.
  \item \textbf{Training Orchestration:} Finally, to complete the ``learning from practice'' loop, the framework channels the vast experiential data generated by the runtime into external training modules, allowing the agent's core policy to be continuously improved.
\end{itemize}

Collectively, these components provide the foundational infrastructure for the entire ``learning from practice'' lifecycle. They empower agents to autonomously interact with complex environments at scale, thereby enabling the efficient synthesis of high-quality experiential data needed for continuous improvement.

\begin{figure}[!t]
  \centering
  \includegraphics[width=0.9\textwidth]{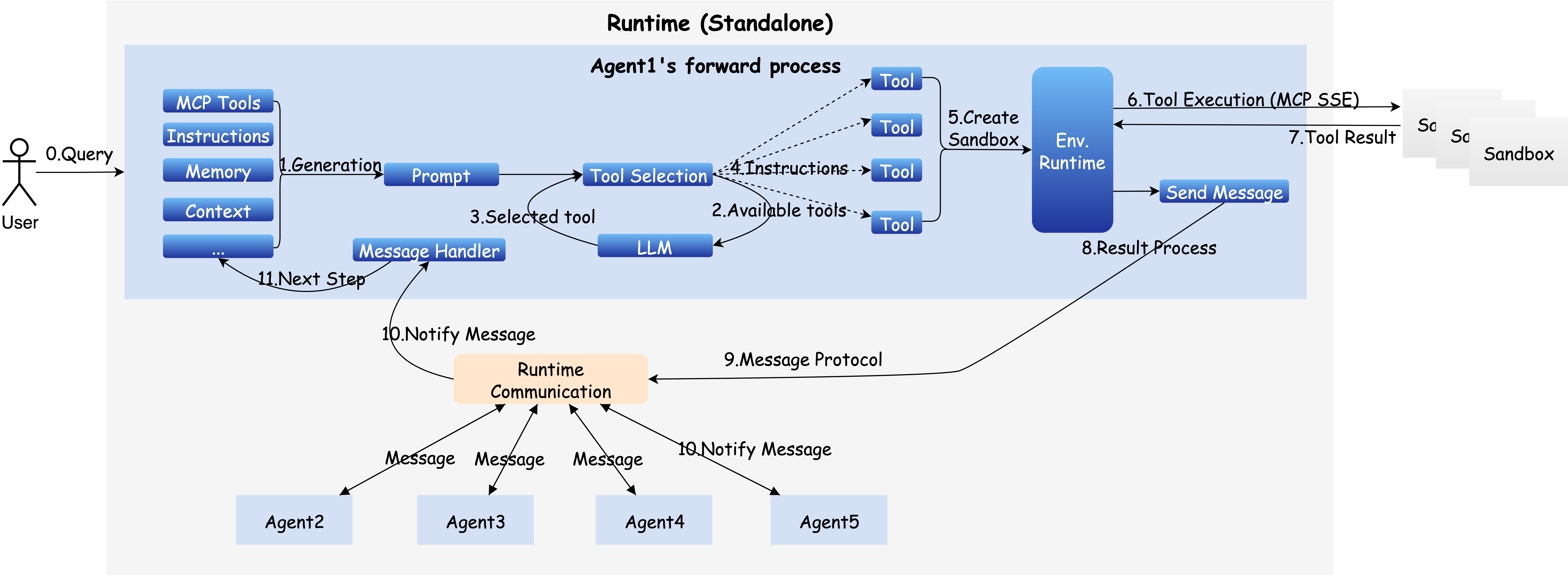}
  \caption{\textbf{An illustration of the runtime in \aworld}, showing the message workflow when an agent receives a query from a user.}
  \label{fig:agent_construction}
\end{figure}

\subsection{Agent Construction}
\label{agent_cons}
As illustrated in~\cref{fig:agent_construction}, the agent architecture consists of several key components that enable flexible and effective task handling.
Upon receiving a user query, the agent first gathers relevant contextual information, including available tools (such as MCP tools), instructions, memory, and other environmental context.
This information is synthesized to generate a prompt.
The agent autonomously selects the most appropriate tool from a diverse pool, which may include built-in functions, externally registered tools via MCP, or even other agents acting as tools. Subsequently, the backend LLM determines the next action to be taken.
The selected action is then executed within an isolated environment that can interact with a sandbox, thereby ensuring secure and reproducible execution.
The results from tool execution are processed by the agent and, if necessary, communicated to other agents or system components through a unified message-passing mechanism.
This event-driven communication enables agents to notify, coordinate, and delegate tasks dynamically, supporting both single-agent and multi-agent workflows.
The agent runtime further supports extensibility through custom message handlers, allowing agents to respond to external events and collaborate on complex operations.

Our framework offers comprehensive support for agent construction, enabling users to flexibly assemble agents for diverse scenarios. Specifically, we provide:
\begin{itemize}[leftmargin=12pt, nosep]
  \item \textbf{Prompt Assembly:} Users can define system prompts to guide agent behavior and tailor responses to specific application scenarios.
  \item \textbf{Custom Toolsets:} The framework allows users to specify the set of environments and tools accessible to each agent, including browser-based interfaces, terminal emulators, and agent-as-tool functionalities.
  \item \textbf{Agent Topology Configuration:} For multi-agent systems, both automated and user-defined custom topologies or workflows are supported, enabling dynamic team formation and tailored collaboration strategies.
\end{itemize}

\subsection{Communication Protocols}
\label{com_pro}
Drawing inspiration from the design of~\citet{googleevents}, \aworld adopts a similar architecture.
As illustrated in~\cref{fig:agent_construction}, \aworld utilizes the \textit{Message} object as the core abstraction to unify three primary communication channels: (1) user-to-agent communication; (2) intra-agent communication between models and tools (e.g., Anthropic’s MCP~\citep{mcp2024}); and (3) inter-agent communication (e.g., Google's A2A protocol~\citep{googlea2a}).
The structure of the \textit{Message} object is defined as follows:

\begin{tcolorbox}[
    title=Message API Specification,
    colback=aworld_blue!10,
    colframe=aworld_blue,
    colbacktitle=aworld_blue,       % 标题背景色
    coltitle=white,
    boxrule=0.5pt,
    arc=2mm,
    fonttitle=\bfseries,
    coltitle=white,
    left=1mm,
    right=1mm,
    top=0.8mm,
    bottom=0.8mm,
    breakable=true,
  ]

  \textbf{Message}\
  \textit{attributes}

  \vspace{1mm}
  \begin{itemize}[leftmargin=1.5em,itemsep=0.3em]
    \item \textbf{id} (str): Unique message identifier (UUID).
    \item \textbf{session\_id} (str): Identifier for the task/session context.
    \item \textbf{sender} (str): The current sender agent/tool.
    \item \textbf{receiver} (Optional[str]): The designated recipient.
    \item \textbf{caller} (Optional[str]): The sender's parent caller in a call chain.
    \item \textbf{payload} (Any): Main content (e.g., ActionModel, Observation, TaskItem).
    \item \textbf{category} (str): Event type or message category.
    \item \textbf{topic} (Optional[str]): Topic-based routing channel (for pub-sub patterns).
    \item \textbf{priority} (int): Execution priority assigned by sender.
    \item \textbf{headers} (Dict[str, Any]): Additional metadata (e.g., task\_id, trace info).
    \item \textbf{timestamp} (float): Epoch timestamp for message creation.
  \end{itemize}

\end{tcolorbox}

The \texttt{payload} field, as the core information carrier, contains standard objects such as \texttt{ActionModel}, \texttt{Observation}, or \texttt{TaskItem}, which are sent by Agents, Tools, and Tasks, respectively. Our communication protocol ensures robust execution through built-in mechanisms for parameter validation, error handling, and result interpretation. For example, if communication with an unavailable or failed agent is attempted, our framework automatically generates an error notification for the sender. This enables systematic exception handling and enhances the overall robustness of distributed task execution.

\subsection{Runtime State Management}
\label{dist_env}

\begin{figure}[!t]
  \centering
  \includegraphics[width=0.95\linewidth]{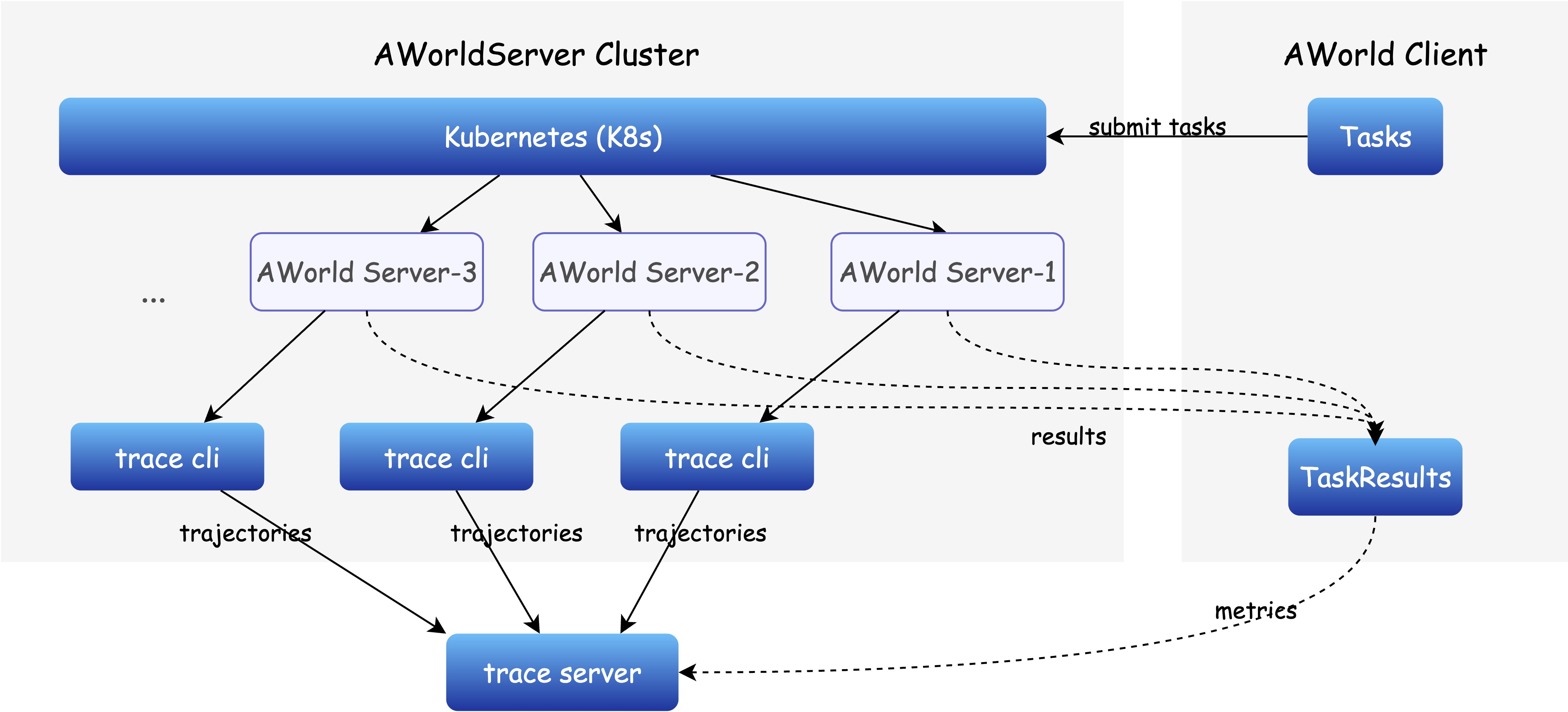}
  \caption{\textbf{Orchestrating Massively Parallel Rollouts in AWorld.}
  The system's distributed architecture, managed by Kubernetes, is engineered to generate vast amounts of training data by concurrently executing agent tasks across numerous, sandboxed environments.}
  \label{fig:AWorld_Distributed}
\end{figure}

To effectively address complex real-world tasks, \aworld adopts a distributed architecture that prioritizes robustness and scalability.
As illustrated in~\cref{fig:AWorld_Distributed}, this architecture is essential for supporting long-horizon agent interactions, with key features including:

\paragraph{High Concurrency.}
To enable agents to accumulate sufficient interactive experience for self-learning, the framework supports high-concurrency execution.
This is achieved through a dynamic task management module powered by Kubernetes, which orchestrates the scheduling, distribution, and prioritization of a large number of concurrent tasks across a distributed cluster of worker nodes.
By efficiently managing heterogeneous workloads and maximizing resource utilization, the system accelerates the generation of rollout samples necessary for both evaluation and continuous agent improvement.

\paragraph{State Consistency.}
State consistency across distributed nodes is maintained via synchronized remote data storage and a centralized trace server, ensuring coherent task execution and rapid recovery from potential disruptions.
The architecture also systematically collects and stores agent trajectories and metrics, further supporting large-scale evaluation and self-improving learning processes.

\subsection{Training Orchestration}
\label{subsec:training_orchestration}

\begin{figure}[!t]
  \centering
  \includegraphics[width=0.9\textwidth]{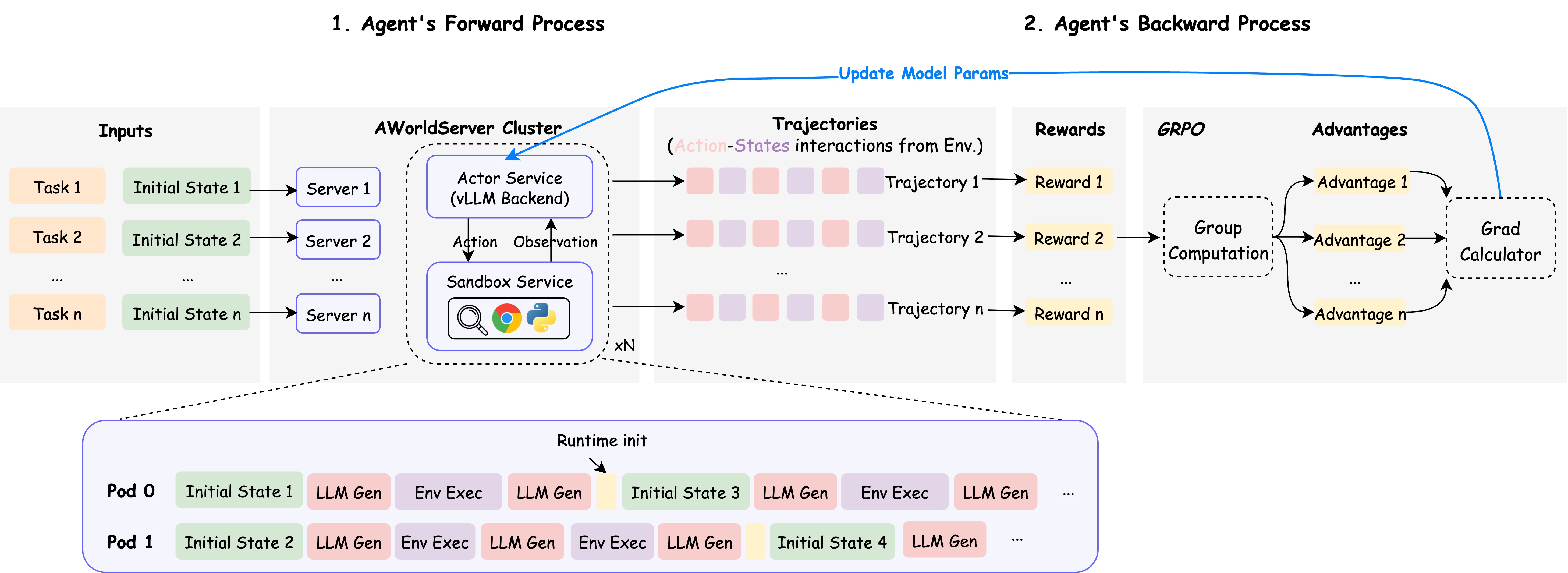}
  \caption{\textbf{An action-state rollout demonstration utilizing \aworld's distributed environments}. \aworld leverages Kubernetes to manage parallel environments, where each environment is encapsulated within a fundamental execution unit known as a \texttt{pod}. In our setup, multiple \texttt{pods} run concurrently across the cluster to enable massive-scale experience generation.}
  \label{fig:training_orchestration}
\end{figure}

There are two commonly used methods to improve the capabilities of the underlying LLM in post-training phase, namely SFT and RL. SFT typically requires high-quality human-labeled data or a complex data synthesis pipeline, which is costly and generally not scalable. In contrast, through RL, agents could learn from environment feedback, which is more accessible and scalable.

A standard RL algorithm, such as GRPO, generally involves three key stages: exploration (rollout), feedback (reward), and learning (gradient update).
Among these, exploration becomes the primary bottleneck when dealing with complex, realistic tasks.
In the case of GAIA tasks, a single rollout can take up to \textit{20 minutes} to complete, significantly slowing down training.
The high-concurrency execution capabilities of \aworld can substantially improve exploration efficiency in such scenarios.

As shown in~\cref{fig:training_orchestration}, \aworld provides a decoupled, high-concurrency execution engine designed to integrate seamlessly with external RL training frameworks such as SWIFT~\citep{zhao2025swift}.
Specifically, the rollout module in conventional frameworks is replaced with the \aworld Executor.
During the \textbf{rollout phase}, tasks are dispatched to the \aworld Executor, which interacts with the inference engine of the RL framework to query actions at each step.
The selected action is executed in the environment, and the corresponding feedback is collected. This interaction continues iteratively, forming a complete trajectory of experience.
Once the trajectory is collected, the training framework assumes control of the \textbf{learning stage}: it performs gradient updates and synchronizes the updated model parameters with the inference engine.

\paragraph{Summary.}
\aworld offers a modular, high-performance framework for building and training intelligent agents. Its general-purpose design accommodates both single-agent and multi-agent settings, enables scalable interaction with realistic environments, and supports seamless integration with external LLMs and RL frameworks.
\section{Experiment}
\label{sec:experiment}

To validate the necessity and effectiveness of the \aworld framework, we conduct a series of experiments. Our evaluation is designed to first establish the relationship between the volume of rollouts and agent performance on complex tasks. Subsequently, we quantify the efficiency gains \aworld provides in generating this experience, demonstrating its critical role in making the ``learning from practice'' paradigm computationally feasible.

\subsection{Experimental Settings}
\label{subsec:exp_settings}

This section outlines the experimental setup, including the benchmark, models, and infrastructure used to validate our claims.

\paragraph{Benchmark.}
All experiments are performed on the GAIA benchmark~\citep{mialon2023gaia}, a challenging testbed for agentic AI that mirrors the complexity of real-world problems.
GAIA's difficulty stems from two primary factors. First, it presents a \textbf{Large Search Space}, characterized by a combinatorial action space of diverse tools and their parameters, a vast observation space filled with variable and often noisy tool outputs, and the necessity for long-horizon reasoning across extended trajectories.
Second, this complexity leads to \textbf{Low Search Efficacy} in current agents, which commonly exhibit suboptimal behaviors such as insufficient planning, performing redundant actions, and demonstrating path dependence without reflecting on past failures.
Collectively, these challenges establish GAIA as an ideal benchmark for rigorously evaluating Agentic AI systems, while also highlighting the critical need for scalable experience generation to overcome its inherent complexity.

\paragraph{Foundation Models.}
Our experiments are centered on training \texttt{Qwen3-32B}~\citep{yang2025qwen3}, a powerful open-source foundation model. To benchmark its performance, we compare our results against several state-of-the-art models. This includes leading closed-source models like \texttt{GPT-4o}~\citep{hurst2024gpt} and \texttt{Claude-3.7-Sonnet}~\citep{anthropic2024claude37sonnet}, as well as another powerful open-source competitor \texttt{DeepSeek-V3}~\citep{liu2024deepseek}.

\paragraph{Hardware Infrastructure.}
Our setup employs a train-inference decoupled architecture to optimize resource utilization for agent training. The training process runs on a dedicated node featuring $8$ NVIDIA A100 ($80$GB) GPUs and a $96$-core CPU. This node is allocated $1200$GB of system memory to support memory-intensive optimization strategies such as DeepSpeed ZeRO3~\citep{rajbhandari2020zero}. A separate, parallel node is dedicated to environment interaction and rollout generation. This inference node is equipped with an identical set of $8$ NVIDIA A100 GPUs and a $96$-core CPU, but is configured with $800$GB of system memory, which is ample for high-throughput agent inference.

\paragraph{Agent Development Framework.}
Our agent development is powered by \aworld, which integrates specialized frameworks to create a seamless pipeline for experience generation and model training. For the rollout phase, \aworld leverages the vLLM~\citep{kwon2023efficient} to manage high-throughput agent inference and interaction with the environment. Subsequently, the collected trajectories are processed by the SWIFT framework~\citep{zhao2025swift}, which orchestrates the model's fine-tuning and reinforcement learning updates. This integrated approach allows us to efficiently manage the entire ``practice-then-learn'' cycle. We will detail the training process in~\cref{sec:gaia_experiment}.

\paragraph{Tool Integrations.}
To equip the agent with a versatile set of capabilities for tackling complex tasks, \aworld integrates a suite of powerful tools, summarized in~\cref{tab:tool_integrations}. These tools provide sandboxed execution environments, web automation, and specialized data processing services.

\begin{table}[h]
  \caption{\textbf{Overview of the integrated tools within the \aworld framework.}}
  \label{tab:tool_integrations}
  \centering
  \resizebox{\textwidth}{!}{%
    \begin{tabular}{ l p{14cm} }
      \toprule
      \textbf{Tool} & \textbf{Functionality} \\
      \midrule
      \texttt{e2b-code-server}\tablefootnote{\url{https://e2b.dev/}} & A sandboxed code instance that compiles and executes arbitrary code snippets. \\
      \texttt{terminal-controller} & Enables terminal command execution, directory navigation, and file system operations. \\
      \texttt{excel} & Lightweight headless Excel engine that reads and writes on \texttt{.xlsx} sheets. \\
      \texttt{calculator} & Basic arithmetic and symbolic expression evaluator. \\
      \texttt{ms-playwright}\tablefootnote{\url{https://github.com/microsoft/playwright}} & Automates browser tasks such as page interaction, web scraping and screenshots.  \\
      \texttt{audio\_server} & On-the-fly audio processing via FFmpeg and Whisper pipelines\tablefootnote{\url{https://github.com/openai/whisper}}. \\
      \texttt{image\_server} & VLM-based image Question \& Answering service powered by models. \\
      \texttt{google-search} & Google Search interface that returns ranked web URLs and summary snippets. \\
      \bottomrule
    \end{tabular}%
  }
\end{table}

\subsection{The Impact of Rollout Scale on Performance}
\label{subsec:rollout_scaling}

\begin{figure}[!t]
  \centering
  \includegraphics[width=0.9\columnwidth]{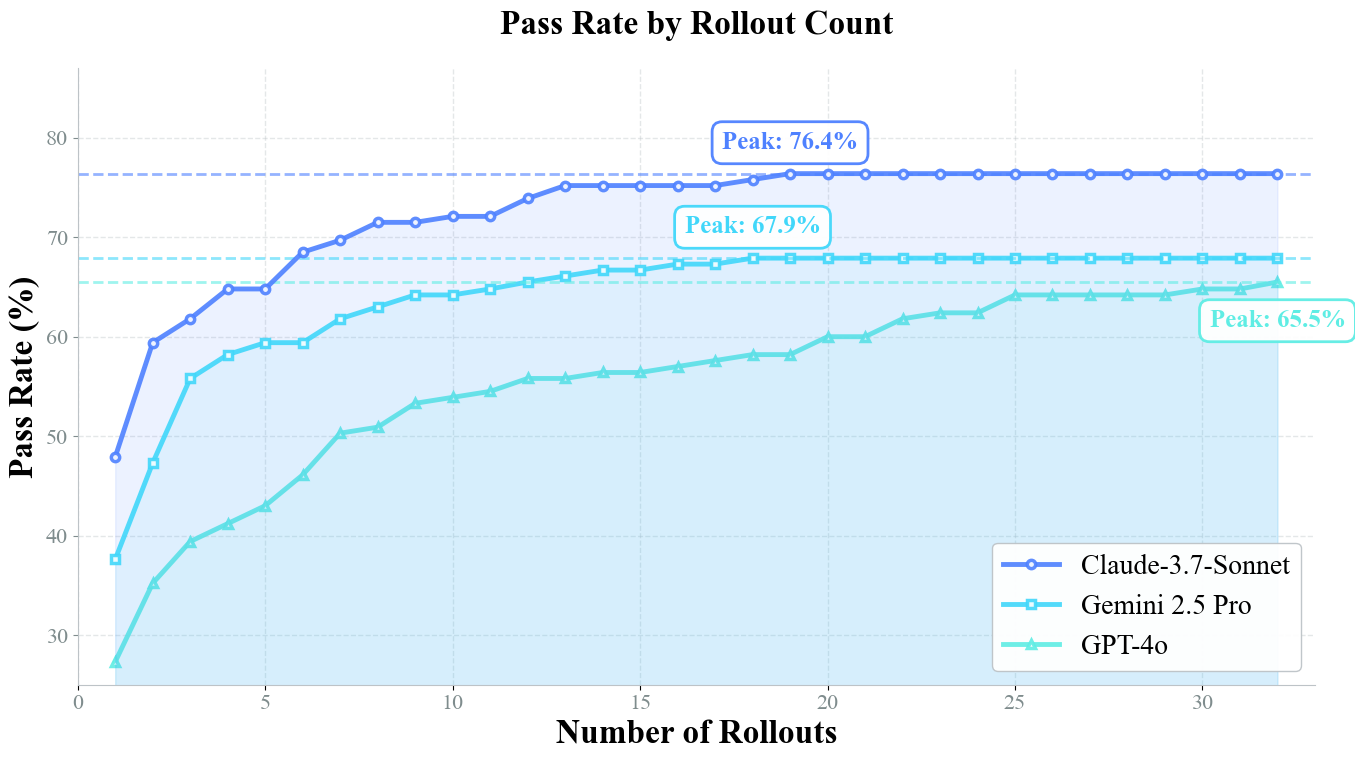}
  \caption{\textbf{Pass Rate as a Function of Rollout Scale on the GAIA Validation Set.} We plot the pass@k success rate for three leading models on the \textbf{full $165$-question GAIA validation set}, varying the number of rollouts (k) from $1$ to $32$. A clear and universal trend emerges: all models, regardless of their initial capability, achieve substantial performance gains with more interaction attempts.}
  \label{fig:rollout_scaling}
\end{figure}

In reinforcement learning, agent improvement hinges on learning from successful examples.
For complex, multi-step tasks like those in GAIA, a single attempt has a low probability of success, making the discovery of these positive reward signals a significant challenge.
To investigate the relationship between interaction budget and problem-solving success, we conducted a comprehensive experiment to quantify the pass rate as a function of the number of rollouts per task. We evaluated three state-of-the-art models—\texttt{Claude-3.7-Sonnet}, \texttt{Gemini 2.5 Pro}~\citep{comanici2025gemini}, and \texttt{GPT-4o}—on the \textbf{entire $165$-question GAIA validation set}, allowing up to $32$ rollouts per question.

As illustrated in~\cref{fig:rollout_scaling}, the results reveal a consistent and crucial trend: increasing the number of rollouts directly and substantially improves the pass rate for all models. For instance, \texttt{Claude-3.7-Sonnet}'s performance climbs from a pass@1 of $47.9\%$ to a peak of $76.4\%$—a gain of nearly $30$ percentage points. Similarly, GPT-4o's success rate more than doubles, rising from $27.3\%$ to $65.5\%$. For most models, the sharpest gains occur within the first $10$-$15$ rollouts, after which performance begins to plateau as they approach their peak problem-solving capacity.

This finding empirically confirms that a sufficient rollout count is essential not merely for data volume, but for ensuring the agent has successful examples to learn from.
Consequently, the efficiency of this rollout process becomes the critical bottleneck for the entire ``learning from practice'' loop.

\subsection{Efficiency of Distributed Rollouts with \aworld}
\label{subsec:efficiency}

\begin{table}[h]
  \centering
  \caption{\textbf{Time comparison for one cycle of rollout and training.} \aworld's distributed executor reduces rollout time by a factor of \textbf{$14.6\times$} compared to a local sequential setup, demonstrating its necessity for scalable agent training.}
  \label{tab:rollout_time}
  \begin{tabular}{lccc}
    \toprule
    \textbf{Rollout Method} & \textbf{Rollout Time (s)} & \textbf{Training Time (s)} & \textbf{Total Time (s)} \\
    \midrule
    \aworld Executor (Distributed) & \colorbox{aworld_blue}{\color{white}$525$} & $144$ & \colorbox{aworld_cyan}{\color{black}$669$} \\
    Sequential Executor (Single-Node) & $7695$ & $144$ & $7839$ \\
    \bottomrule
  \end{tabular}%
\end{table}

Given the established need for large-scale data generation, this section evaluates the core efficiency of \aworld. We measure the wall-clock time for a full cycle of experience generation and model training, comparing our distributed approach against a standard single-node setup.

One might ask why the single-node baseline is sequential rather than parallel. A naive parallel implementation on a single node is \textit{not a viable baseline} for complex environments like GAIA. The combination of resource-intensive tools (e.g., a full browser engine) and long-horizon tasks creates significant CPU and memory demands. Attempting to run multiple such rollouts concurrently on one machine leads to severe \textbf{resource contention} and \textbf{process instability}, making sequential execution the \textit{only stable and practical configuration} for a single-node setup and a fair point of comparison.

As shown in \cref{tab:rollout_time}, the efficiency gains are substantial. The \aworld Executor completes the rollout phase in just \textbf{$525$ seconds}, while the Sequential Executor requires \textbf{$7695$ seconds}. This translates to a \textbf{$14.6$-fold speedup} in experience generation. Crucially, as the training time ($144$s) is constant, the total cycle time is reduced from $7839$ seconds to a mere \textbf{$669$ seconds}.
This result provides clear quantitative evidence for a central thesis of our work: for complex agentic tasks, the primary bottleneck has shifted from \textit{training computation} to \textit{environmental interaction}. \aworld is purpose-built to dismantle this bottleneck, making it an indispensable tool for the scalable ``learning from practice'' paradigm.

\subsection{Training an Agent with \aworld: Performance on the GAIA Benchmark}
\label{sec:gaia_experiment}
To validate the framework's effectiveness for both evaluation and training, this section presents our training experiments and performance on benchmarks.

\paragraph{Implementations.}

To provide the agent with a strong initial policy and mitigate the cold-start problem, we first perform SFT.
For this initial training phase, we curate a dataset of $886$ successful trajectories, sampled using the Claude 3.7 Sonnet model.
This SFT-trained model then serves as the starting point for our reinforcement learning process.
The subsequent reinforcement learning loop is composed of three key steps, as shown in~\cref{fig:training_orchestration}:

\begin{enumerate}[leftmargin=12pt, nosep]
  \item \textbf{Rollout}. Tasks are submitted to the \aworld Executor. During each action step, the \aworld Executor queries the vLLM module of SWIFT to generate an action. The action is then executed in the environment, and the corresponding feedback is collected. This process iterates over multiple steps to produce a complete trajectory. In particular, the model weights underlying the vLLM module are updated at each training step. For each task, we set the number of rollouts to $32$.

  \item \textbf{Reward Calculator.} We employ a rule-based reward mechanism, where the agent receives a reward of $1$ if its generated answer exactly matches the ground truth, and $0$ otherwise.

  \item \textbf{Gradient Update.} Following the approach in~\citet{shao2024deepseekmath}, we employ the GRPO algorithm for advantage estimation and gradient update computation. These updates are executed within the SWIFT framework, after which the updated model is synchronized with the vLLM server.
\end{enumerate}

\begin{table*}[!t]
  \centering
  \caption{\textbf{Performance comparison on the GAIA test set (pass@1) and xbench-DeepSearch(pass@1).} Our RL-enhanced model, Qwen3-32B-\aworld, demonstrates significant improvements over the base model. Scores are reported in percentage (\%). Best results are in \textbf{bold}, second-best are \underline{underlined}.}
  \label{tab:gaia_results}
  \begin{tabular}{l cccc c}
    \toprule
    \multirow{2}{*}{\textbf{Model}} & \multicolumn{4}{c}{\textbf{GAIA}} & \multirow{2}{*}{\textbf{
        \begin{tabular}[c]{@{}c@{}}xbench-\\DeepSearch
    \end{tabular}}} \\
    \cmidrule(lr){2-5}
    & \textbf{Avg. (\%)} & \textbf{Level 1 (\%)} & \textbf{Level 2 (\%)} & \textbf{Level 3 (\%)} & \\
    \midrule
    GPT-4o & $27.91$ & $40.86$ & $24.53$ & \underline{$14.29$} & $30$ \\
    Claude 3.7 Sonnet & \textbf{$43.85$} & \textbf{$64.52$} & \textbf{$40.88$} & \underline{$14.29$} & \textbf{$45$} \\
    DeepSeek-V3 & $31.89$  & \underline{$52.69$}  & $25.16$  & $14.29$ & \underline{$35$} \\
    Qwen3-32B (Base) & $21.59$ & $30.11$ & $22.01$ & $4.08$ & $12$ \\
    \rowcolor{aworld_blue!50}
    Qwen3-32B-\aworld & \underline{$32.23$} \textcolor{green!50}{(+$10.6\%$)} & $47.31$ \textcolor{green!50}{(+$17.2\%$)} & \underline{$28.30$} \textcolor{green!50}{(+$6.3\%$)} & \textbf{$16.33$} \textcolor{green!50}{(+$12.3\%$)} & $32$ \textcolor{green!50}{(+$20.0\%$)} \\
    \bottomrule
  \end{tabular}
\end{table*}

\paragraph{Result Analysis.}

The main performance results of our trained agent on the GAIA test set, supplemented by the xbench-DeepSearch benchmark~\citep{chen2025xbench}, are presented in \cref{tab:gaia_results}.
The analysis reveals two key findings. First, the reinforcement learning process, enabled by the \aworld framework, yields substantial performance gains over the base Qwen3-32B model.
The overall pass@1 accuracy on GAIA improves by $10.6$ absolute percentage points (from $21.59\%$ to $32.23\%$), with significant improvements observed across all difficulty levels.
Furthermore, the strong performance on xbench-DeepSearch—improving from $12\%$ to $32\%$ \textit{without any direct training} on its samples—indicates that the agent has acquired robust, generalizable problem-solving skills rather than overfitting to the GAIA environment.

Second, when compared with state-of-the-art models, our Qwen3-32B-\aworld agent demonstrates highly competitive performance. Its overall GAIA score is comparable to DeepSeek-V3 and surpasses GPT-4o. Notably, the most significant achievement is observed on the benchmark's most challenging Level 3 questions. Here, our agent achieves a pass@1 score of $16.33\%$, outperforming all other listed models, including leading proprietary systems. This specific result highlights the effectiveness of our training methodology in enhancing complex, multi-step reasoning capabilities.
\section{Future work}

Building on the foundation of \aworld, our future work will focus on advancing towards collective and self-improving intelligence. Our roadmap consists of three main stages.
First, we will extend the framework to support the deployment of \textbf{multi-agent systems} in diverse and heterogeneous environments, enabling collaborative problem-solving.
Second, we aim to cultivate \textbf{specialized, capability-centric agents} that achieve expert-level performance in distinct domains such as complex reasoning or web navigation, forming a society of experts.
Ultimately, our goal is to enable these systems to achieve a degree of \textbf{autonomous self-improvement}, where agents learn continuously from their collective practice to refine not only their skills but also their collaboration strategies, creating a truly self-sustaining learning loop.

\section{Conclusion}

In this paper, we introduced \aworld, an open-source framework designed to realize the ``learning from practice'' paradigm for Agentic AI. We established that for complex tasks, exemplified by the challenging GAIA benchmark, the primary obstacle to this paradigm is the inefficiency of experience generation. \aworld tackles this bottleneck directly, leveraging a distributed architecture to achieve a \textbf{14.6-fold speedup} in data collection. This efficiency enabled us to train a Qwen3-32B-based agent that not only significantly surpasses its base model but also delivers highly competitive performance against leading proprietary models. Our work provides both a practical infrastructure and empirical validation for the ``learning from practice'' paradigm, paving the way for the development of more capable and self-improving agents.

%%%%%%%%%%%%%%%%%%%%%%%%%%%%%%%%%%%%%%%%%%%%%%%%%%%%%%%%%%%%%%%%%%
%%%%%%%%%%%%%%%%%%%%%%%%%%%%%%%%%%%%%%%%%%%%%%%%%%%%%%%%%%%%%%%%%%
%%%%%%%%%%%%%%%%%%%%%%%%%%%%%%%%%%%%%%%%%%%%%%%%%%%%%%%%%%%%%%%%%

% reference
\clearpage
\bibliographystyle{config/antgroup}
\bibliography{references}

% appendix
% \clearpage
% \appendix
% \input{doc/appendix}

\end{document}